\documentclass[conference]{IEEEtran}
\IEEEoverridecommandlockouts
\usepackage{cite}
\usepackage{amsmath,amssymb,amsfonts}
\usepackage{algorithmic}
\usepackage{graphicx}
\usepackage{float}
\usepackage{amsmath}
\usepackage{subcaption}
\usepackage{textcomp}
\usepackage{xcolor}
\def\BibTeX{{\rm B\kern-.05em{\sc i\kern-.025em b}\kern-.08em
    T\kern-.1667em\lower.7ex\hbox{E}\kern-.125emX}}
\begin{document}

\title{Motion planning in high-dimensional spaces}

\author{\IEEEauthorblockN{Luka Petrovi\'{c}}
\IEEEauthorblockA{University of Zagreb Faculty of Electrical Engineering and Computing, \\
Department of Control and Computer Engineering,\\
Laboratory for Autonomous Systems and Mobile Robotics (LAMOR).  \\
Email: luka.petrovic@fer.hr}
}

\maketitle

\begin{abstract}
Motion planning is a key tool that allows robots to navigate through an environment without collisions.
The problem of robot motion planning has been studied in great detail over the last several decades, with researchers initially focusing on systems such as planar mobile robots and low degree-of-freedom (DOF) robotic arms. 
The increased use of high DOF robots that must perform tasks in real time in complex dynamic environments spurs the need for fast motion planning algorithms. 
In this overview, we discuss several types of strategies for motion planning in high dimensional spaces and dissect some of them, namely grid search based, sampling based and trajectory optimization based approaches.
We compare them and outline their advantages and disadvantages, and finally, provide an insight into future research opportunities.
\end{abstract}


\section{Introduction}
A fundamental need in robotics is to have algorithms that convert high-level specifications of tasks into low level descriptions of how to move.
The term motion planning is often used for these kinds of problems \cite{lavalle2006planning}. 
Motion planning is therefore an indispensable skill for robots that aspire to navigate through an environment without collisions.
Motion planning algorithms attempt to generate trajectories through the robot's configuration space that are both feasible and optimal based on some performance criterion that may vary depending on the task, robot, or environment.
There are two common paradigms for successful motion planning.
The first paradigm separates motion planning into path planning and path execution.
The second paradigm directly finds the trajectory of the robot - it takes into account dynamic constraints to plan the full robot motion from one point to the other.
Path planning is used to find the shortest, or otherwise optimal, collision-free path between two points in environment while not taking into account the temporal component of robot motion.
Path execution is then used to follow the planned path, leveraging methods from control theory.
The initial attempts to solve path planning problems included grid search methods such as the A$^*$ algorithm \cite{hart1968formal}, reactive planners such as bug algorithms \cite{ng2007performance}, combinatorial planning methods originating from computational geometry such as cell decompositions \cite{latombe2012robot}, visibility graphs \cite{lozano1979algorithm} and Voronoi diagrams~\cite{o1985retraction, takahashi1989motion}.

All of the mentioned methods are complete, meaning that they find a solution if it exists and report failure otherwise.
They present elegant solutions for low dimensional configuration spaces with static obstacles.
However, bug algorithms produce unnecessarily long paths, while grid search methods and combinatorial approaches suffer from the so-called curse of dimensionality, i.e. they quickly become computationally intractable with the increase of the configuration space dimension.
That also means that replanning is possible, making those approaches ineffective in dynamic environments.

The increasing complexity of robots and the environments that they operate in has spurred the need for high-dimensional motion planning.
Consider, for instance, a personal robot operating in a cluttered household environment or a humanoid robot performing navigation and manipulation tasks in an unstructured environment.
Efficient motion planning is important to enable these high degree-of-freedom (DOF) robots to perform tasks, subject to motion constraints while avoiding collisions with obstacles in the environment.
Processing time is especially important in dynamic environments where replanning is necessary.
Those considerations lead to the development of grid-based methods with ameliorated efficiency and to the development of sampling-based motion planning algorithms which offer weaker guarantees than combinatorial methods but are more efficient.
Sampling-based algorithms abandon the concept of explicitly charaterizing the configuration space - they use a collision detection algorithm to probe the configuration space to see whether some configuration lies in free space or not.
Sampling-based algorithms are probabilistically complete, meaning that the probability they will produce a solution approaches one as more time is spent.

All of the discussed techniques so far aim at capturing the connectivity of free configuration space into a graph.
The exigency for efficient and fast planning methods lead to development of the planning paradigm which directly finds the trajectory of the robot. 
Potential field methods were the initial approach of that paradigm.
They model the robot, which is represented as a point in configuration space, as a particle under the influence of an artificial potential field.
The artificial potential field is created by the attractive force from goal point and repulsive forces from obstacles in configuration space.
To obtain the policy which moves the robot safely to the goal, one would simply perform gradient descent on the potential function.
Main disadvantages of potential field methods are the lack of completness and optimality guarantees and the presence of local minima for a non point-mass robot.

Recently, the trajectory optimization methods that are appropriate for very high DOF robots have been proposed.
The trajectory is encoded as a sequence of states and controls.
Trajectory optimization approaches start with an initial trajectory and then minimize an objective function in order to optimize the trajectory according to some criterion.
Those approaches are exceptionally fast, however, unlike sampling-based approaches, they only find locally optimal solutions.

The overview is organized as follows. 
Section \ref{sec:grid} elaborates grid-based approaches to motion planning. 
In Section \ref{sec:sampling}, sampling based planning methods are presented.
Afterwords, Section \ref{sec:trajopt} describes trajectory optimization methods.
Lastly, conclusion is given in Section \ref{sec:conclusion} with focus on importance of high-dimensional motion planning.
\section{Grid-based approaches}
\label{sec:grid}
Grid-based motion planning approaches overlay a grid on configuration space and assume each configuration is corresponding to a grid point.
The robot is allowed to move to adjacent grid points as long as the line between them is collision-free, i.e. contained within free configuration space.
This discretizes the set of actions, and search algorithms, for instance the A$^*$ algorithm \cite{hart1968formal}, are used to find a path from the start to the goal.
These approaches require setting a grid resolution.
Search is faster with lower resolution, but the algorithm will fail to find paths through narrow portions of free configuration space.
The D$^*$ algorithm \cite{stentz1994optimal}, along with its improved variants Focussed D$^*$ \cite{stentz1995focussed} and D$^*$ Lite \cite{koenig2005fast}, is a commonly used algorithm based on the A$^*$ capable of planning paths in unknown, partially known, and changing environments in an optimal and complete manner.

Altough naive application of A$^*$ is typically unsuitable for high-dimensional planning problems, since number of points on the grid grows exponentially in the configuration space dimension, current research has made advances in applying forward search to systems with many DOFs.
Until recently, a major problem with A$^*$ and related algorithms had been that admissible heuristics result in examination of prohibitively large portions of the configuration space, whereas inflated heuristics cause significantly suboptimal behavior.
In \cite{likhachev2004ara}, a framework for efficiently updating an A$^*$ search while smoothly reducing heuristic inflation is presented, allowing resolution complete search in an anytime fashion on a broader variety of problems than previously computable.
It starts by finding a suboptimal solution quickly using a loose bound, then tightens the bound progressively as time allows.
Given enough time it finds a provably optimal solution. 
Additional work has examined inducing smoother motion while reducing the cardinality of the action set \cite{cohen2010search}, as well as reusing partial plans discovered by past searches \cite{phillips2012graphs}.
Since grid-based methods are complete, they are often used for planning in environments featuring bottlenecks and other such narrow passages.
The main drawback of grid-based approaches is their computational complexity, as even the state-of-the-art methods suffer from the curse of dimensionality, and become computationally intractable for very high DOF systems such as humanoid robots.
\section{Sampling-based methods}
\label{sec:sampling}
Sampling-based approaches have become popular in the domain of high-dimensional motion planning, including manipulation planning.
In a sense, these approaches attempt to capture the connectivity of the robot configuration space by sampling it \cite{elbanhawi2014sampling}.
Randomized approaches have its advantages in terms of providing efficient solutions for challenging problems \cite{lavalle2006planning}.
The downside is that the solutions are widely regarded as suboptimal.
Sampling based planners are understood to be probabilistically complete \cite{lavalle2000rapidly}, a weaker notion of completeness that ensures a solution will be provided, if one exists, given sufficient runtime of the algorithm (even if it means infinite runtime).
However, they cannot determine if no solution exists.
\begin{figure}
	\vspace{0.5\baselineskip}
	\centering
	\begin{subfigure}[]{0.49\columnwidth}
		\includegraphics[width=\textwidth]{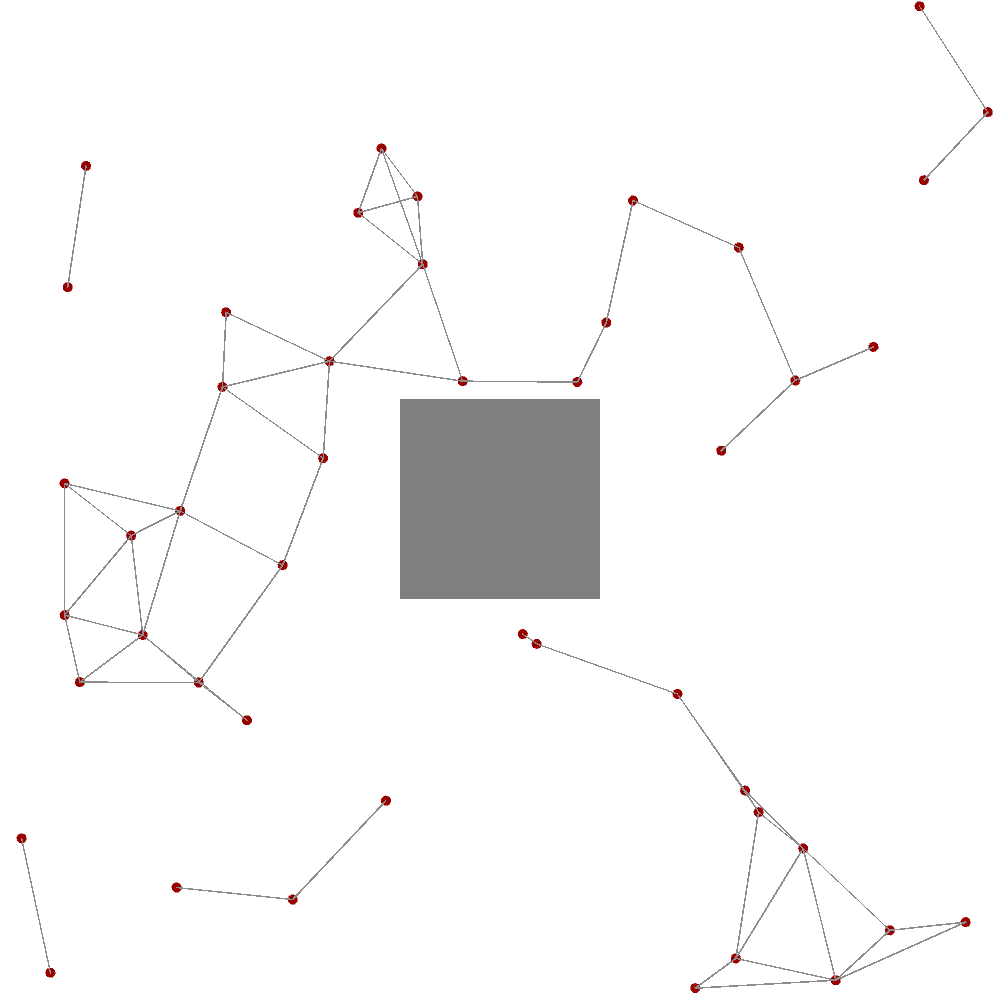}
	\end{subfigure}
	\begin{subfigure}[]{0.49\columnwidth}
		\includegraphics[width=\textwidth]{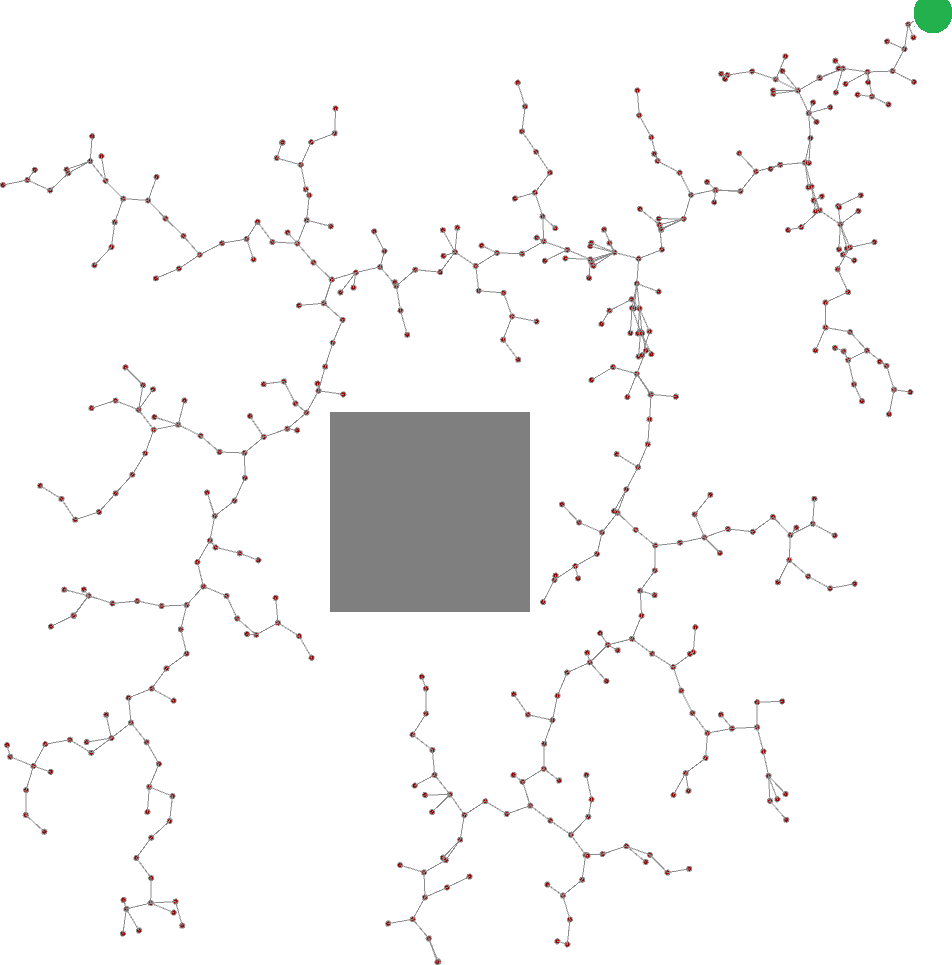}
	\end{subfigure}
	\vspace{1mm}
	\caption{Left: roadmap built in the PRM learning phase. Right: RRT exploring environment with one obstacle after 500 iterations \cite{elbanhawi2014sampling}. The root of the tree is shown as green circle. }
	\label{fig:sampling}
\end{figure}
Sampling-based approaches typically work in a two-step fashion.
First, a collision-free path is discovered without regard for any measure of cost.
Second, the obtained path is improved by applying certain heuristics.
For the first step, a Monte-Carlo algorithm for path planning with many degrees of freedom \cite{barraquand1990monte} was a seminal work that demonstrated a solver for impressively difficult problems.
Perhaps the most commonly used algorithms are the probabilistic roadmap (PRM) \cite{kavraki1996probabilistic} and rapidly exploring random trees (RRT) \cite{kuffner2000rrt}.
The PRM method was shown to be well suited for path planning in configuration spaces with many DOFs, and with complex constraints, including kinodynamic \cite{kuffner1999autonomous, hsu2000randomized}. 
RRT has also been applied to differential constraints, and was shown to be successful for general high dimensional planning \cite{lavalle2001randomized}.
The intuitive implementation of both RRT and PRM, and the quality of the solutions, lead to their widespread adoption in robotics and many other fields.
Even though the idea of connecting points sampled randomly from the state space is essential in both RRT and PRM, these two algorithms differ in the way that they construct a graph connecting these points \cite{karaman2011sampling}.
Other sampling-based path planners of note include expansive space trees (EST) \cite{hsu1997path} and sampling-based roadmap of trees (SRT) \cite{plaku2005sampling}.
The EST employs a function that sets the probability of node selection based on neighboring nodes, unlike RRT where sampling is uniform.
The SRT combines the main features of multiple-query algorithms such as PRM with those of single-query algorithms such as EST and RRT.
For improving the planned path, one popular approach is the \textit{shortcut} heuristic, which picks pairs of configurations along the path and invokes a local planner to attempt to replace the intervening sub-path with a shorter one \cite{chen1998sandros, kavraki1998probabilistic}.
Methods such as medial axis and partial shortcuts have also proven effective \cite{geraerts2006creating}.

PRM implements two main procedures to generate a probabilistic roadmap.
A learning phase occurs first, where the configuration space is randomly sampled for a certain amount of time.
The sampled configurations are declared as vertices if they lie in free configuration space and are connected to nearby vertices with a local planner, while those in the obstacle space are discarded.
This is followed by a query phase where the start and goal configurations are defined and connected to the roadmap.
A graph search is then performed to find the shortest path through the roadmap between start and goal configurations.
An example of a roadmap built in the PRM learning phase is depicted in Fig. \ref{fig:sampling}. 
Roadmaps are sometimes referred to as forests, as an analogy to trees in RRT.
PRM has the inherent ability to solve different instances of the planning problem in the same environment, which is a product of maintaining the roadmap and specifying start and goal configurations in a subsequent stage.
It is therefore referred to as a multi-query planner.
Planning time is invested in sampling and generating a roadmap, but queries are solved quickly.
PRM was initially developed for articulated robots, but has been extended for non-holonomic car-like robots.

RRT represents another category of sampling based planners, which are single-query planners.
A tree is incrementally grown from the start configuration to the goal configuration, or vice versa.
A configuration is randomly selected in the configuration space.
If it lies in the free space, a connection is attempted to the nearest vertex in the tree.
As a result of uniform sampling, the RRT is more likely to select samples in larger Voronoi regions and the tree is incrementally and rapidly grown towards that free space \cite{elbanhawi2014sampling}.
An example of a tree generated by RRT algorithm exploring environment with one obstacle is shown in Fig. \ref{fig:sampling}.
For single query problems, RRT is faster compared to PRM, since it does not require a learning phase.

Sampling based approaches typically do not explicitly optimize an objective function, altough variants of PRM and RRT which are provably asymptotically optimal have been proposed in \cite{karaman2011sampling}.
As the sampling-based planners became increasingly well understood in recent years, it was suggested that randomization may not, by itself, account for their efficiency \cite{lavalle2004relationship}.
It was shown that quasi-random sampling sequences can accomplish similar or better performance than their randomized counterparts \cite{branicky2001quasi}.
The main disadvantage of sampling-based motion planning methods is that the obtained paths often manifest redundant and jerky motion and hence require post processing to smooth and shorten the computed trajectories.
Furthermore, considerable computational effort is expended in sampling and connecting samples in portions of the configuration space that might not be relevant to the task.
\section{Trajectory optimization}
\label{sec:trajopt}
The key motivation for trajectory optimization is the focus on producing \textit{optimal motion}: incorporating dynamics, smoothness, and obstacle avoidance in a mathematically precise objective.
Despite a rich theorethical history and successful applications, most notably in the control of spacecraft and rockets, trajectory optimization techniques have had limited success in motion planning.
Much of this may be attributed to two causes: the large computational cost fot evaluating objective functions and their higher-order derivatives in high-dimensional spaces, and the presence of local minima when considering motion planning as a (generally non-convex) continuous optimization problem.

A significant amount of recent work has focused on trajectory optimization and related problems.
The use of potential fields for avoiding obstacles, including dynamic ones, was first proposed in \cite{khatib1986real}.
The method’s sensitivity to local minima has been addressed in a range of related work.
Analytical navigation functions that are free of local minima have been proposed for some specific environments \cite{rimon1992exact}.
A global potential field to push the robot away from configuration space obstacles, starting with a trajectory that was in collision was proposed in \cite{warren1989global}.
As a part of a local optimization that tries to shorten and smooth the trajectory, the free part of configuration space can be locally approximated as a union of spheres around the current trajectory \cite{quinlan1993elastic}.
The trajectory is modelled as a mass-spring system, an elastic band, and replanning is performed by scanning back and forth along the elastic, while moving one mass particle at a time.
An extensive effort is made to construct a model of the free space, and the cost function contains terms that attempt to control the motion of the trajectory particles along the elastic.
This method was further extended in \cite{brock2002elastic} where the real-time application to a high-degree-of-freedom humanoid robot was presented.
The aforementioned approaches locally approximate the free space using a union of spheres, which is an overly conservative approximation and may not find feasible trajectories even if they exist.

\begin{figure}
	\centering
	\begin{subfigure}[]{0.49\columnwidth}
		\includegraphics[width=\textwidth]{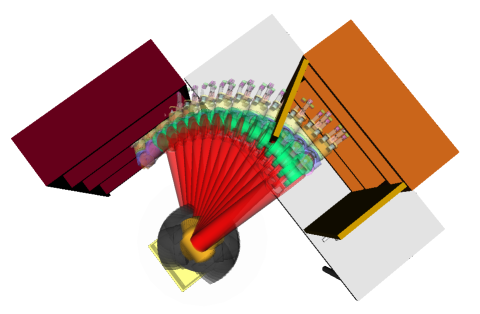}
	\end{subfigure}
	\begin{subfigure}[]{0.49\columnwidth}
		\includegraphics[width=\textwidth]{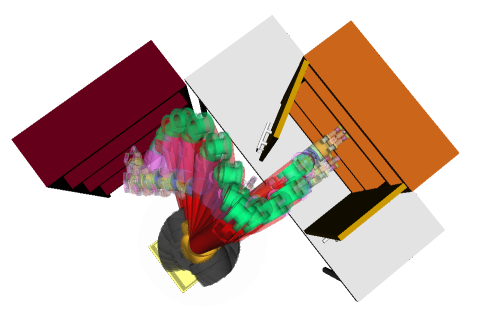}
	\end{subfigure}
	\caption{Left: the initial straight-line trajectory through robot's configuration space. Right: the final trajectory post optimization \cite{chomp-ijrr}}
	\label{fig:traj}
\end{figure}

More recent trajectory optimization methods play two important roles in robot motion planning.
Firstly, they can be used to smooth and shorten trajectories computed by other planning methods such as sampling-based planners.
Secondly, they can be used to compute locally optimal, collision-free trajectories from scratch starting from naive trajectory initializations that might be in collision with obstacles \cite{schulman2014motion}.
The state-of-the-art trajectory optimization methods all start with an initial (commonly straight-line) trajectory and then minimize an objective function in order to optimize the trajectory according to some criterion.
An illustrative example of motion planning for mobile manipulator in simulation is shown in Fig. \ref{fig:traj}.
The figure shows both the initial trajectory that is in collision with obstacles and the optimized, collision-free one. 
The goal of trajectory optimization for motion planning is to obtain trajectories $\theta (t)$ that minimize costs and satisfy constraints.
Motion planning can thus be formalized as
\begin{equation}
\begin{aligned}
& \text{minimize} 
&& \mathcal{F} [\theta (t)] \\
& \text{subject to} 
&& \mathcal{G}_i [\theta (t)] \leq 0, i = 1, \dots,  m_{ineq} \\
&&& \mathcal{H}_i [\theta (t)] = 0, i = 1, \dots,  m_{eq}
\end{aligned}
\label{eq:traj}
\end{equation}
where the trajectory $\theta (t)$ is a continuous-time function that maps time $t$ to states, which are usually robot configurations and perchance higher order derivatives.
$\mathcal{F} [\theta (t)]$ is a cost functional or objective that encodes collision costs that enforce the trajectory to be \textit{collision-free} and the quality of a trajectory that usually corresponds to \textit{smoothness} and is evaluated as a minimization of velocity or acceleration \cite{chomp-ijrr}. 
$\mathcal{G}_i [\theta (t)]$ represents inequality constraints, for example joint angle limits, and $\mathcal{H}_i [\theta (t)]$ represents task-dependent equality constraints, such as the desired start and end configurations and velocities or the desired end-effector orientation, i.e. holding a glass full of liquid upright.
Depending on the specific problem, the number of inequality constraints may be zero.
Collision cost may also appear as an obstacle avoidance inequality constraint \cite{schulman2014motion}, depending on the optimization technique used to solve \eqref{eq:traj}.
Most existing trajectory optimization algorithms in practice work with a finely discretized trajectory, which is good for reasoning about tight navigation constraints or thin obstacles, but can cause a large computational cost.

\begin{figure}
	\vspace{0.7\baselineskip}
	\centering
	\begin{subfigure}[]{0.49\columnwidth}
		\includegraphics[width=\textwidth]{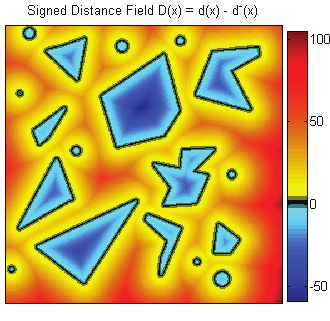}
	\end{subfigure}
	\begin{subfigure}[]{0.49\columnwidth}
		\includegraphics[width=\textwidth]{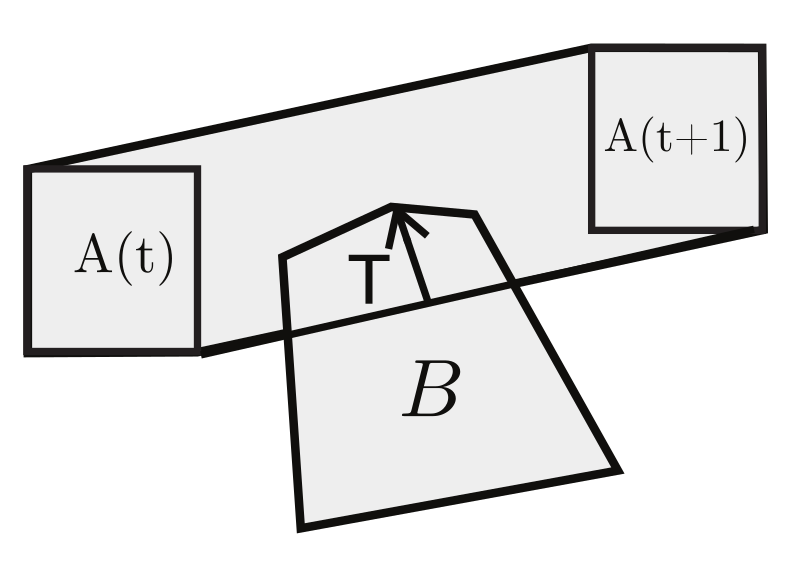}
	\end{subfigure}
	\vspace{2mm}
	\caption{Left: signed distance field for use in collision cost in CHOMP \cite{chomp-ijrr} and GPMP2 \cite{gpmp2}. Right: illustration of swept volume for use in continuous collision cost in TrajOpt \cite{schulman2014motion}}
	\label{fig:traj2}
\end{figure}

Covariant Hamiltonian Optimization for Motion Planning (CHOMP) \cite{chomp}, \cite{chomp-ijrr} revived interest in trajectory optimization methods by demonstrating the effectiveness on several robotic platforms, including a mobile manipulation platform and a quadruped.
The key feature of CHOMP is a formulation of trajectory costs that are invariant to the time parametrization of the trajectory.
In order to produce smooth robot motion while avoiding obstacles, CHOMP uses two objective functionals: a smoothness term which captures dynamics of the trajectory and an obstacle functional which captures the requirement of avoiding obstacles and preferring margin from them.
They define the smoothness functional in terms of a metric in the space of trajectories. 
The obstacle functional is developed as a line integral of a scalar cost field $c$ - a precomputed signed distance field, defined so that it is invariant to retiming.
Regardless of how fast or slow the arm moves through the field, it will accumulate the exact same cost.
An example of a precomputed signed distance field used for the obstacle functional is shown in Fig. \ref{fig:traj2}.
The two functionals have complementary role: the obstacle functional governs the shape of the path, and the smoothness functional governs the timing along the path.

Several augmentations of CHOMP have been proposed.
Multigrid CHOMP with local smoothing \cite{he2013multigrid} which improves the runtime of CHOMP under constraints, whithout significantly reducing optimality.
T-CHOMP \cite{byravan2014space} is a functional gradient algorithm that directly optimizes in space-time, thus being able to successfully incorporate constraints and cost functions that explicitly depend on time. 
Inceremental trajectory optimization algorithm (ITOMP) \cite{park2012itomp} enables real-time replanning in dynamic environments.
In contrast to CHOMP, which exploits the availability of the gradient, Stochastic Trajectory Optimization for Motion Planning (STOMP) \cite{stomp} samples a series of noisy trajectories to explore the space around an initial trajectory which are then combined to produce an updated trajectory with lower cost.
The key trait of STOMP is its ability to optimize non-differentiable constraints.

An important shortcoming of CHOMP and related methods is the need for many trajectory states in order to reason about fine resolution obstacle representations or find feasible solutions when there are many constraints.
The framework called TrajOpt \cite{trajopt}, \cite{schulman2014motion} formulates motion planning as sequential quadratic programming problem and features convex collision checking.
Sequential convex optimization involves solving a series of convex optimization problems that approximate the cost and constraints of the original problem.
The ability to add new constraints and costs to the optimization problem allowes TrajOpt to tackle a larger range of motion planning problems, including planning for underactuated, non-holonomic systems.
For collisions, TrajOpt uses signed distances using convex-convex collision detection and takes into account continuous-time safety by considering the swept-out volume of the robot between time steps. An illustration of swept volume is shown in Fig. \ref{fig:traj2}.
This formulation allows for sparsely sampled trajectory, which makes TrajOpt very computationally efficient in practice.
However, if the smoothness is required in the output trajectory, either a densely parametrized trajectory or post-processing of the trajectory might still be needed thus increasing computation time.

A continous-time trajectory representation can avoid the computational cost incurred by using large number of states and yield a more efficient approach, while maintaining the smoothness of trajectory.
Linear interpolation, splines, and hierarchical wavelets have been used to represent trajectories in state estimation and filtering. Recently, kernel methods \cite{Marinho-RSS-16} and B-splines \cite{elbanhawi2016randomized} have been used in a similar manner to represent trajectories with fewer states in motion planning problems.

The Gaussian process motion planning family of algorithms \cite{gpmp, gpmp2, gpmpgraph, gpmp-ijrr, gpmplie} uses continous-time trajectory representation; specifically, they view trajectories as functions that map time to robot state.
The continuous-time trajectory is represented as a sample from a Gaussian process (GP) generated by a linear time-varying stochastic differential equation.
They show that GPs inherently provide a notion of trajectory optimality through a \textit{prior}.
Efficient structure-exploiting GP regression facillitates querying the trajectory at any time of interest in $\mathcal{O}(1)$ complexity \cite{barfoot2014batch}.
Using this representation, they developed GPMP (Gaussian Process Motion Planner), a gradient-based optimization algorithm that can efficiently overcome the large computational costs of fine discretization while maintaining smoothness in the result.

Through the GP formulation, they view motion planning as probabilistic inference \cite{Toussaint2009, Toussaint2010}.
Similarly to the notion of trajectory optimality being captured by a prior on trajectories, one can also view the notion of feasibility probabilistically, encoded in a likelihood function.
Through the use of factor graphs \cite{kschischang2001factor}, Bayesian inference can be used to compute a solution to the motion planning problem efficiently.
The duality between optimization and inference allows performing efficient inference on factor graphs, thus exploiting the structure of the underlying system by solving sparse least squares problems.
Similar approaches have been used to solve Simultanous Localization and Mapping (SLAM) problems \cite{dellaert2006square}.
With this key insight, preexisting efficient optimization methods developed by the SLAM community can be exploited for motion planning.
The GPMP2 algorithm, which is more efficient than previous motion planning algorithms, stems from those considerations.
A useful property of GPMP2 is its extensibility and applicability for wide range of problems.
For example, combined learning from demonstration and motion planning \cite{rana2017towards} presented an efficient approach to skill learning and generalizable skill reproduction.
In \cite{maric2018singularity}, authors provided a framework for avoidance of singular robot configurations as manipulability maximization, while a unified probabilistic framework for trajectory estimation and planning was provided in \cite{steap}.
The main drawback of GPMP algorithms is that they are limited in their ability to work with nonlinear inequality constraints.
\section{Conclusion and future work}
\label{sec:conclusion}
Safely navigating through an environment is one of the key tasks which an autonomous mobile robot or vehicle has to accomplish.
Motion planning is an essential tool used to find trajectories of robot states that achieve a desired task.
In this overview, we have presented some of the representative methods for high-dimensional motion planning.
Grid-based approaches are resolution complete and often offer optimal solutions.
However, the number of grid points grows exponentially in the configuration space dimension, which makes even the state-of-the-art methods inappropriate for very high-dimensional problems.
Sampling-based approaches are efficient in most practical problems but offer weaker guarantees.
They are probabilistically complete, however, they often require post-processing and can still be inefficient in very complex configuration spaces.
Trajectory optimization approaches can solve high-dimensional motion planning problems quickly, but solutions are only locally optimal.

Grid-based and sampling-based approaches have been well-studied. Most of the future research opportunities for high-dimensional motion planning lie in the most recent area - trajectory optimization.
One of the challenges is finding the right cost function to optimize.
With current methods, it is possible for a trajectory that is in collision briefly but generally stays far away from obstacles to have lower cost than a trajectory that never collides.
Other important challenge is finding a principled way to tackle the local minima problem, either by employing different optimization methods, or by investigating the effect of trajectory initialization.
Sampling-based and grid-search based methods could be used to provide a coarse initial trajectory which is then enhanced with trajectory optimization methods.
Seeking a coherent way to handle constraints and guarantee safety is another open problem.
\bibliographystyle{IEEEtran}
\bibliography{bibliography}

\begin{thebibliography}{10}
\providecommand{\url}[1]{#1}
\csname url@samestyle\endcsname
\providecommand{\newblock}{\relax}
\providecommand{\bibinfo}[2]{#2}
\providecommand{\BIBentrySTDinterwordspacing}{\spaceskip=0pt\relax}
\providecommand{\BIBentryALTinterwordstretchfactor}{4}
\providecommand{\BIBentryALTinterwordspacing}{\spaceskip=\fontdimen2\font plus
\BIBentryALTinterwordstretchfactor\fontdimen3\font minus
  \fontdimen4\font\relax}
\providecommand{\BIBforeignlanguage}[2]{{%
\expandafter\ifx\csname l@#1\endcsname\relax
\typeout{** WARNING: IEEEtran.bst: No hyphenation pattern has been}%
\typeout{** loaded for the language `#1'. Using the pattern for}%
\typeout{** the default language instead.}%
\else
\language=\csname l@#1\endcsname
\fi
#2}}
\providecommand{\BIBdecl}{\relax}
\BIBdecl

\bibitem{lavalle2006planning}
S.~M. LaValle, \emph{Planning algorithms}.\hskip 1em plus 0.5em minus
  0.4em\relax Cambridge university press, 2006.

\bibitem{hart1968formal}
P.~E. Hart, N.~J. Nilsson, and B.~Raphael, ``A formal basis for the heuristic
  determination of minimum cost paths,'' \emph{IEEE transactions on Systems
  Science and Cybernetics}, vol.~4, no.~2, pp. 100--107, 1968.

\bibitem{ng2007performance}
J.~Ng and T.~Br{\"a}unl, ``Performance comparison of bug navigation
  algorithms,'' \emph{Journal of Intelligent and Robotic Systems}, vol.~50,
  no.~1, pp. 73--84, 2007.

\bibitem{latombe2012robot}
J.-C. Latombe, \emph{Robot motion planning}.\hskip 1em plus 0.5em minus
  0.4em\relax Springer Science \& Business Media, 2012, vol. 124.

\bibitem{lozano1979algorithm}
T.~Lozano-P{\'e}rez and M.~A. Wesley, ``An algorithm for planning
  collision-free paths among polyhedral obstacles,'' \emph{Communications of
  the ACM}, vol.~22, no.~10, pp. 560--570, 1979.

\bibitem{o1985retraction}
C.~{\'O}'D{\'u}nlaing and C.~K. Yap, ``A “retraction” method for planning
  the motion of a disc,'' \emph{Journal of Algorithms}, vol.~6, no.~1, pp.
  104--111, 1985.

\bibitem{takahashi1989motion}
O.~Takahashi and R.~J. Schilling, ``Motion planning in a plane using
  generalized voronoi diagrams,'' \emph{IEEE Transactions on robotics and
  automation}, vol.~5, no.~2, pp. 143--150, 1989.

\bibitem{stentz1994optimal}
A.~Stentz, ``Optimal and efficient path planning for partially-known
  environments,'' in \emph{Robotics and Automation, 1994. Proceedings., 1994
  IEEE International Conference on}.\hskip 1em plus 0.5em minus 0.4em\relax
  IEEE, 1994, pp. 3310--3317.

\bibitem{stentz1995focussed}
A.~Stentz \emph{et~al.}, ``The focussed d\^{}* algorithm for real-time
  replanning,'' in \emph{IJCAI}, vol.~95, 1995, pp. 1652--1659.

\bibitem{koenig2005fast}
S.~Koenig and M.~Likhachev, ``Fast replanning for navigation in unknown
  terrain,'' \emph{IEEE Transactions on Robotics}, vol.~21, no.~3, pp.
  354--363, 2005.

\bibitem{likhachev2004ara}
M.~Likhachev, G.~J. Gordon, and S.~Thrun, ``Ara*: Anytime a* with provable
  bounds on sub-optimality,'' in \emph{Advances in neural information
  processing systems}, 2004, pp. 767--774.

\bibitem{cohen2010search}
B.~J. Cohen, S.~Chitta, and M.~Likhachev, ``Search-based planning for
  manipulation with motion primitives,'' in \emph{Robotics and Automation
  (ICRA), 2010 IEEE International Conference on}.\hskip 1em plus 0.5em minus
  0.4em\relax IEEE, 2010, pp. 2902--2908.

\bibitem{phillips2012graphs}
M.~Phillips, B.~J. Cohen, S.~Chitta, and M.~Likhachev, ``E-graphs:
  Bootstrapping planning with experience graphs.'' in \emph{Robotics: Science
  and Systems}, vol.~5, no.~1, 2012.

\bibitem{elbanhawi2014sampling}
M.~Elbanhawi and M.~Simic, ``Sampling-based robot motion planning: A review,''
  \emph{Ieee access}, vol.~2, pp. 56--77, 2014.

\bibitem{lavalle2000rapidly}
S.~M. LaValle and J.~J. Kuffner~Jr, ``Rapidly-exploring random trees: Progress
  and prospects,'' 2000.

\bibitem{barraquand1990monte}
J.~Barraquand and J.-C. Latombe, ``A monte-carlo algorithm for path planning
  with many degrees of freedom,'' in \emph{Robotics and Automation, 1990.
  Proceedings., 1990 IEEE International Conference on}.\hskip 1em plus 0.5em
  minus 0.4em\relax IEEE, 1990, pp. 1712--1717.

\bibitem{kavraki1996probabilistic}
L.~E. Kavraki, P.~Svestka, J.-C. Latombe, and M.~H. Overmars, ``Probabilistic
  roadmaps for path planning in high-dimensional configuration spaces,''
  \emph{IEEE transactions on Robotics and Automation}, vol.~12, no.~4, pp.
  566--580, 1996.

\bibitem{kuffner2000rrt}
J.~J. Kuffner and S.~M. LaValle, ``Rrt-connect: An efficient approach to
  single-query path planning,'' in \emph{Robotics and Automation, 2000.
  Proceedings. ICRA'00. IEEE International Conference on}, vol.~2.\hskip 1em
  plus 0.5em minus 0.4em\relax IEEE, 2000, pp. 995--1001.

\bibitem{kuffner1999autonomous}
J.~J. Kuffner, ``Autonomous agents for real-time animation,'' Ph.D.
  dissertation, PhD thesis, Stanford University, 1999.

\bibitem{hsu2000randomized}
D.~Hsu, \emph{Randomized single-query motion planning in expansive
  spaces}.\hskip 1em plus 0.5em minus 0.4em\relax Stanford University USA,
  2000.

\bibitem{lavalle2001randomized}
S.~M. LaValle and J.~J. Kuffner~Jr, ``Randomized kinodynamic planning,''
  \emph{The international journal of robotics research}, vol.~20, no.~5, pp.
  378--400, 2001.

\bibitem{karaman2011sampling}
S.~Karaman and E.~Frazzoli, ``Sampling-based algorithms for optimal motion
  planning,'' \emph{The international journal of robotics research}, vol.~30,
  no.~7, pp. 846--894, 2011.

\bibitem{hsu1997path}
D.~Hsu, J.-C. Latombe, and R.~Motwani, ``Path planning in expansive
  configuration spaces,'' in \emph{Robotics and Automation, 1997. Proceedings.,
  1997 IEEE International Conference on}, vol.~3.\hskip 1em plus 0.5em minus
  0.4em\relax IEEE, 1997, pp. 2719--2726.

\bibitem{plaku2005sampling}
E.~Plaku, K.~E. Bekris, B.~Y. Chen, A.~M. Ladd, and L.~E. Kavraki,
  ``Sampling-based roadmap of trees for parallel motion planning,'' \emph{IEEE
  Transactions on Robotics}, vol.~21, no.~4, pp. 597--608, 2005.

\bibitem{chen1998sandros}
P.~C. Chen and Y.~K. Hwang, ``Sandros: a dynamic graph search algorithm for
  motion planning,'' \emph{IEEE Transactions on Robotics and Automation},
  vol.~14, no.~3, pp. 390--403, 1998.

\bibitem{kavraki1998probabilistic}
L.~E. Kavraki and J.-C. Latombe, ``Probabilistic roadmaps for robot path
  planning,'' 1998.

\bibitem{geraerts2006creating}
R.~Geraerts and M.~H. Overmars, ``Creating high-quality roadmaps for motion
  planning in virtual environments,'' in \emph{Intelligent Robots and Systems,
  2006 IEEE/RSJ International Conference on}.\hskip 1em plus 0.5em minus
  0.4em\relax IEEE, 2006, pp. 4355--4361.

\bibitem{lavalle2004relationship}
S.~M. LaValle, M.~S. Branicky, and S.~R. Lindemann, ``On the relationship
  between classical grid search and probabilistic roadmaps,'' \emph{The
  International Journal of Robotics Research}, vol.~23, no. 7-8, pp. 673--692,
  2004.

\bibitem{branicky2001quasi}
M.~S. Branicky, S.~M. LaValle, K.~Olson, and L.~Yang, ``Quasi-randomized path
  planning,'' in \emph{Robotics and Automation, 2001. Proceedings 2001 ICRA.
  IEEE International Conference on}, vol.~2.\hskip 1em plus 0.5em minus
  0.4em\relax IEEE, 2001, pp. 1481--1487.

\bibitem{khatib1986real}
O.~Khatib, ``Real-time obstacle avoidance for manipulators and mobile robots,''
  in \emph{Autonomous robot vehicles}.\hskip 1em plus 0.5em minus 0.4em\relax
  Springer, 1986, pp. 396--404.

\bibitem{rimon1992exact}
E.~Rimon and D.~E. Koditschek, ``Exact robot navigation using artificial
  potential functions,'' \emph{IEEE Transactions on robotics and automation},
  vol.~8, no.~5, pp. 501--518, 1992.

\bibitem{warren1989global}
C.~W. Warren, ``Global path planning using artificial potential fields,'' in
  \emph{Robotics and Automation, 1989. Proceedings., 1989 IEEE International
  Conference on}.\hskip 1em plus 0.5em minus 0.4em\relax IEEE, 1989, pp.
  316--321.

\bibitem{quinlan1993elastic}
S.~Quinlan and O.~Khatib, ``Elastic bands: Connecting path planning and
  control,'' in \emph{Robotics and Automation, 1993. Proceedings., 1993 IEEE
  International Conference on}.\hskip 1em plus 0.5em minus 0.4em\relax IEEE,
  1993, pp. 802--807.

\bibitem{brock2002elastic}
O.~Brock and O.~Khatib, ``Elastic strips: A framework for motion generation in
  human environments,'' \emph{The International Journal of Robotics Research},
  vol.~21, no.~12, pp. 1031--1052, 2002.

\bibitem{schulman2014motion}
J.~Schulman, Y.~Duan, J.~Ho, A.~Lee, I.~Awwal, H.~Bradlow, J.~Pan, S.~Patil,
  K.~Goldberg, and P.~Abbeel, ``Motion planning with sequential convex
  optimization and convex collision checking,'' \emph{The International Journal
  of Robotics Research}, vol.~33, no.~9, pp. 1251--1270, 2014.

\bibitem{chomp-ijrr}
M.~Zucker, N.~Ratliff, A.~D. Dragan, M.~Pivtoraiko, M.~Klingensmith, C.~M.
  Dellin, J.~A. Bagnell, and S.~S. Srinivasa, ``Chomp: Covariant hamiltonian
  optimization for motion planning,'' \emph{The International Journal of
  Robotics Research}, vol.~32, no. 9-10, pp. 1164--1193, 2013.

\bibitem{chomp}
N.~Ratliff, M.~Zucker, J.~A. Bagnell, and S.~Srinivasa, ``Chomp: Gradient
  optimization techniques for efficient motion planning,'' in \emph{Robotics
  and Automation, 2009. ICRA'09. IEEE International Conference on}, 2009, pp.
  489--494.

\bibitem{he2013multigrid}
K.~He, E.~Martin, and M.~Zucker, ``Multigrid chomp with local smoothing,'' in
  \emph{Humanoid Robots (Humanoids), 2013 13th IEEE-RAS International
  Conference on}.\hskip 1em plus 0.5em minus 0.4em\relax IEEE, 2013, pp.
  315--322.

\bibitem{byravan2014space}
A.~Byravan, B.~Boots, S.~S. Srinivasa, and D.~Fox, ``Space-time functional
  gradient optimization for motion planning,'' in \emph{Robotics and Automation
  (ICRA), 2014 IEEE International Conference on}.\hskip 1em plus 0.5em minus
  0.4em\relax IEEE, 2014, pp. 6499--6506.

\bibitem{park2012itomp}
C.~Park, J.~Pan, and D.~Manocha, ``Itomp: Incremental trajectory optimization
  for real-time replanning in dynamic environments.'' in \emph{ICAPS}, 2012.

\bibitem{stomp}
M.~Kalakrishnan, S.~Chitta, E.~Theodorou, P.~Pastor, and S.~Schaal, ``Stomp:
  Stochastic trajectory optimization for motion planning,'' in \emph{Robotics
  and Automation (ICRA), 2011 IEEE International Conference on}, 2011, pp.
  4569--4574.

\bibitem{trajopt}
J.~Schulman, J.~Ho, A.~X. Lee, I.~Awwal, H.~Bradlow, and P.~Abbeel, ``Finding
  locally optimal, collision-free trajectories with sequential convex
  optimization.'' in \emph{Robotics: science and systems}, vol.~9, no.~1, 2013,
  pp. 1--10.

\bibitem{Marinho-RSS-16}
Z.~Marinho, B.~Boots, A.~Dragan, A.~Byravan, G.~J. Gordon, and S.~Srinivasa,
  ``Functional gradient motion planning in reproducing kernel hilbert spaces,''
  in \emph{Proceedings of Robotics: Science and Systems}, June 2016.

\bibitem{elbanhawi2016randomized}
M.~Elbanhawi, M.~Simic, and R.~Jazar, ``Randomized bidirectional b-spline
  parameterization motion planning,'' \emph{IEEE Transactions on intelligent
  transportation systems}, vol.~17, no.~2, pp. 406--419, 2016.

\bibitem{gpmp}
M.~Mukadam, X.~Yan, and B.~Boots, ``Gaussian process motion planning,'' in
  \emph{Robotics and Automation (ICRA), 2016 IEEE International Conference on},
  2016, pp. 9--15.

\bibitem{gpmp2}
J.~Dong, M.~Mukadam, F.~Dellaert, and B.~Boots, ``Motion planning as
  probabilistic inference using gaussian processes and factor graphs,'' in
  \emph{Proceedings of Robotics: Science and Systems (RSS-2016)}, 2016.

\bibitem{gpmpgraph}
E.~Huang, M.~Mukadam, Z.~Liu, and B.~Boots, ``Motion planning with graph-based
  trajectories and gaussian process inference,'' in \emph{Robotics and
  Automation (ICRA), 2017 IEEE International Conference on}, 2017, pp.
  5591--5598.

\bibitem{gpmp-ijrr}
M.~Mukadam, J.~Dong, X.~Yan, F.~Dellaert, and B.~Boots, ``Continuous-time
  gaussian process motion planning via probabilistic inference,'' \emph{arXiv
  preprint arXiv:1707.07383}, 2017.

\bibitem{gpmplie}
\BIBentryALTinterwordspacing
J.~Dong, B.~Boots, and F.~Dellaert, ``Sparse gaussian processes for
  continuous-time trajectory estimation on matrix lie groups,'' \emph{Arxiv},
  vol. abs/1705.06020, 2017. [Online]. Available:
  \url{http://arxiv.org/abs/1705.06020}
\BIBentrySTDinterwordspacing

\bibitem{barfoot2014batch}
T.~D. Barfoot, C.~H. Tong, and S.~S{\"a}rkk{\"a}, ``Batch continuous-time
  trajectory estimation as exactly sparse gaussian process regression.'' in
  \emph{Robotics: Science and Systems}, 2014.

\bibitem{Toussaint2009}
\BIBentryALTinterwordspacing
M.~Toussaint, ``Robot trajectory optimization using approximate inference,'' in
  \emph{Proceedings of the 26th Annual International Conference on Machine
  Learning}, ser. ICML '09.\hskip 1em plus 0.5em minus 0.4em\relax New York,
  NY, USA: ACM, 2009, pp. 1049--1056. [Online]. Available:
  \url{http://doi.acm.org/10.1145/1553374.1553508}
\BIBentrySTDinterwordspacing

\bibitem{Toussaint2010}
M.~Toussaint and C.~Goerick, \emph{A Bayesian View on Motor Control and
  Planning}.\hskip 1em plus 0.5em minus 0.4em\relax Springer Berlin Heidelberg,
  2010, pp. 227--252.

\bibitem{kschischang2001factor}
F.~R. Kschischang, B.~J. Frey, and H.-A. Loeliger, ``Factor graphs and the
  sum-product algorithm,'' \emph{IEEE Transactions on information theory},
  vol.~47, no.~2, pp. 498--519, 2001.

\bibitem{dellaert2006square}
F.~Dellaert and M.~Kaess, ``Square root sam: Simultaneous localization and
  mapping via square root information smoothing,'' \emph{The International
  Journal of Robotics Research}, vol.~25, no.~12, pp. 1181--1203, 2006.

\bibitem{rana2017towards}
M.~A. Rana, M.~Mukadam, S.~R. Ahmadzadeh, S.~Chernova, and B.~Boots, ``Towards
  robust skill generalization: Unifying learning from demonstration and motion
  planning,'' in \emph{Conference on Robot Learning}, 2017, pp. 109--118.

\bibitem{maric2018singularity}
F.~Mari{\'c}, O.~Limoyo, L.~Petrovi{\'c}, I.~Petrovi{\'c}, and J.~Kelly,
  ``Singularity avoidance as manipulability maximization using continuous time
  gaussian processes,'' \emph{arXiv preprint arXiv:1803.09493}, 2018.

\bibitem{steap}
M.~Mukadam, J.~Dong, F.~Dellaert, and B.~Boots, ``Simultaneous trajectory
  estimation and planning via probabilistic inference,'' in \emph{Proceedings
  of Robotics: Science and Systems (RSS)}, 2017.

\end{thebibliography}
\end{document}